%% file: Exploring_mining_attributed_streams.tex
\documentclass[sigconf]{acmart}
\usepackage{amsmath}
\usepackage{booktabs} 
\usepackage{caption} 
\usepackage{subcaption} 
\usepackage{graphicx}
\usepackage{pgfplots}
\usepackage[all]{nowidow}
\usepackage[utf8]{inputenc}
\usepackage{tikz}
\usetikzlibrary{er,positioning,bayesnet}
\usepackage{multicol}
\usepackage{algpseudocode,algorithm,algorithmicx}
\usepackage{hyperref}
\usepackage{booktabs}
\usepackage{multirow}
\usepackage{bm}
\usepackage{dsfont}
\usepackage[inline]{enumitem} 

\newlength\Colsep
\begin{document}
\title{Exploring and mining attributed sequences of interactions}
%
%
\author{Tiphaine Viard} 
\affiliation{
\institution{LTCI, Institut Polytechnique de Paris}
\city{Paris}
\country{France}
}
\author{Henry Soldano}
\affiliation{
\institution{NukkAI}
\city{Paris}
\country{France}
}
\affiliation{
\institution{LIPN}
\city{Villetaneuse}
\country{France}
}
\author{Guillaume Santini}
\affiliation{
\institution{LIPN}
\city{Villetaneuse}
\country{France}
}

\renewcommand{\shortauthors}{Viard, Soldano, Santini}
\begin{abstract}
We are faced with data comprised of entities interacting over time: this can be individuals meeting, customers buying products, machines exchanging packets on the IP network, among others.
Capturing the dynamics as well as the structure of these interactions is of crucial importance for analysis.
These interactions can almost always be labeled with content: group belonging, reviews of products, abstracts, etc. We model these stream of interactions as stream graphs, a recent framework to model interactions over time.
Formal Concept Analysis provides a framework for analyzing concepts evolving within a context.
Considering graphs as the context, it has recently been applied to perform closed pattern mining on social graphs. In this paper, we are interested in pattern mining in sequences of interactions.
After recalling and extending notions from formal concept analysis on graphs to stream graphs, we introduce algorithms to enumerate closed patterns on a labeled stream graph, and introduce a way to select relevant closed patterns.
We run experiments on two real-world datasets of interactions among students and citations between authors, and show both the feasibility and the relevance of our method.
\end{abstract}

\maketitle

\input{sections/intro}

\input{sections/rw}

\input{henry}

\section{Stream graphs}
\label{sec:streams}

\begin{figure}
    \centering
    \includegraphics[width=0.48\linewidth]{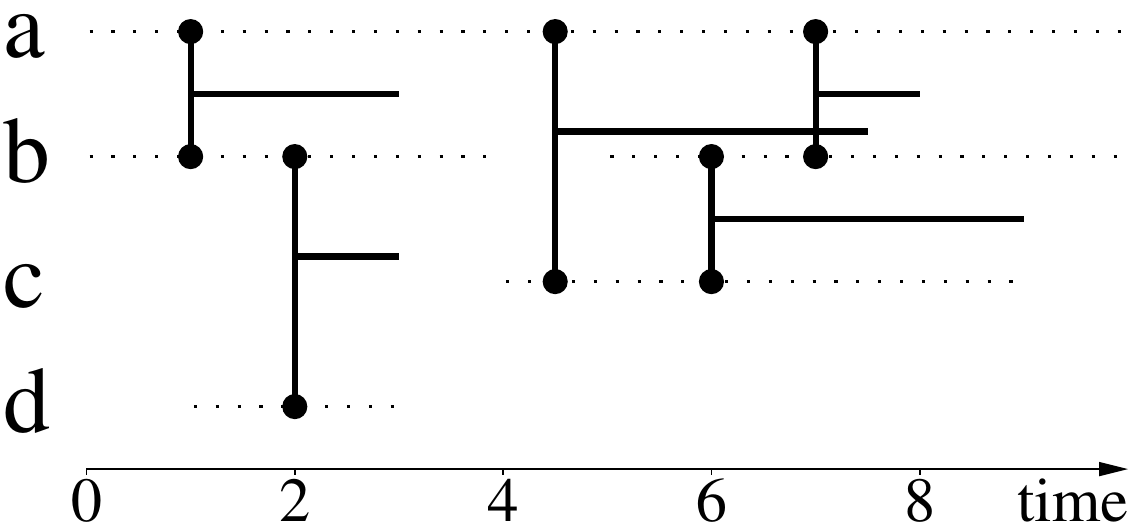}
    \includegraphics[width=0.48\linewidth]{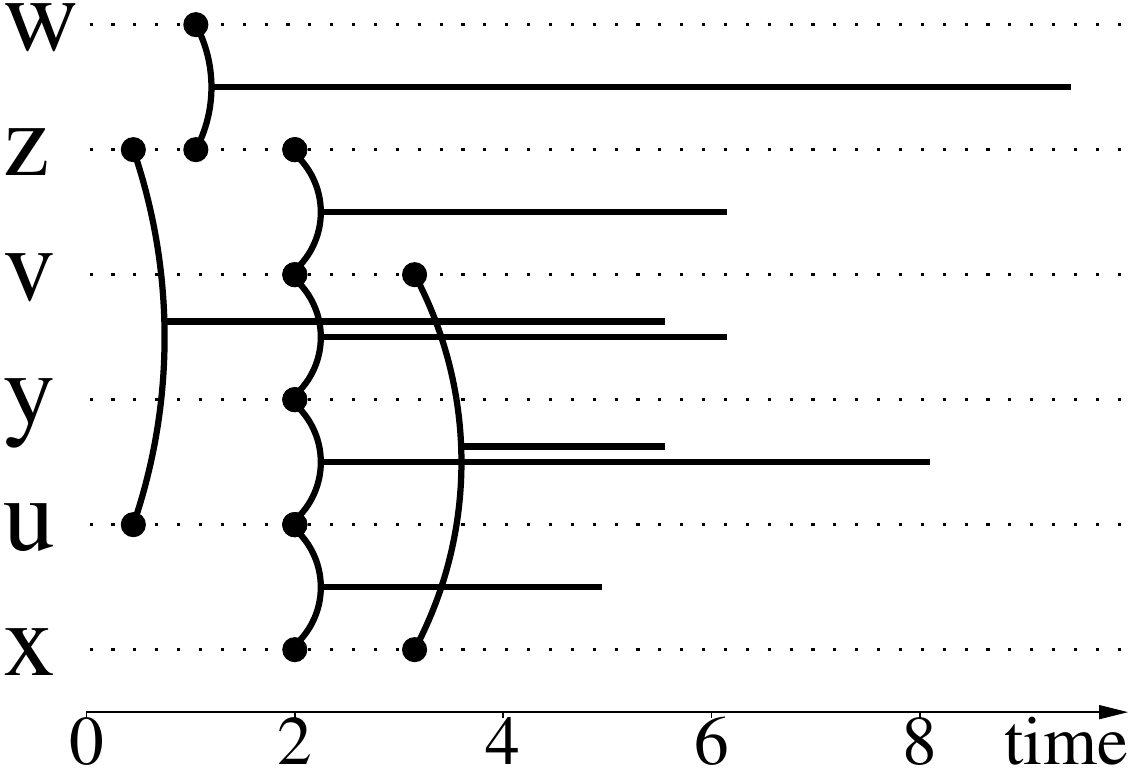}
    \caption{Two toy stream graphs, modelling interactions over $T=[0,10]$. {\bf Left:} A unipartite stream graph involving $4$ nodes $V=\{a,b,c,d\}$ and $W= \{\{a\}\times [0,10], \{b\}\times [0,4]\cup [5,10],
    \{c\}\times [4,10], \{d\}\times [1,3]\}$ and the set of interactions $E=\{\{ab\}\times [1,3]\cup [7,8], \{bd\}\times [2,3], \dots\}$. {\bf Right:} A bipartite stream graph involving $6$ nodes, with $\top=\{u,v, w\}$ and $\bot=\{x,y,z\}$.}
    \label{fig:stream}
\end{figure}

Stream graphs are a recent formalism~\cite{latapy2018stream} to model interactions over time by generalizing many useful notions from complex and social networks analysis.
We denote a {\em stream graph} by the tuple $S=(T,V,W,E)$, where $T$ is a time interval, $V$ a set of nodes.
$W\subseteq T\times V$ denotes the presence times of nodes, such that $(t,v)\in W$ means that node $v$ is "active" at time $t$, and finally, $E\subseteq T\times V\otimes V$ denotes interactions, such that $(t, uv)\in E$ means that nodes $u$ and $v$ interacted at time $t$.
If we consider that interactions are undirected ($(t,uv) = (t,vu)$) and without loop ($u\ne v$) and we denote by $V\otimes V$ the set of such pairs of nodes.
In the directed case, we denote edges as $(t,u,v)\in E$, and $E\subseteq T \times V \times V$.
Figure~\ref{fig:stream} depicts toy stream graphs.

Furthermore, we say that $S' = (T', V', W', E')$ is a substream of $S$ if and only if $T'\subseteq T$, $V'\subseteq V$, $W'\subseteq W$ and $E'\subseteq E$. 
We denote this by $S'\subseteq S$. We denote by $S(W^\prime)$ the substream graph induced by a time-node vertex subset $ W^\prime \subseteq W$, and whose interaction subset $E_{W^\prime}$  contains interaction between time-nodes of $W^\prime$.

Finally,  let us define $G_S = (V_S, E_S)$ the graph induced by $S$, with $V_S = \{u: \exists (t,uv) \in E, t\in T, v\in V \}$ and $E_S = \{uv : \exists (t,uv) \in E, t\in T \}$. In other words, nodes and edges belong to $V_S$ and $E_S$ if and only if there exist some time $t$ such that $(t,uv)$ belongs to $E$.
The adaptation to the directed case is straightforward.

For any node $v\in V$, we denote its \emph{neighbourhood at time $t$} by ${\mathcal N}_t(v) = \{(t,u) : \exists (t,uv)\in E, u \in V\}$ the set of $(t,u)$ that interact with node $v$ at time $t$.
We further denote the {\em degree} of $v$ at time $t$ by $d_t(v) = |{\mathcal N}_t(v)|$.
For example, in Figure~\ref{fig:stream} (left), node $b$ at time $2$ interacts with nodes $a$ and $d$, and so ${\mathcal N}_2(b) = \{a, d\}$, and $d_2(b)= 2$.

We can extend the stream graph definition to directed case, in which  all interactions in $E$ are directed.  In that case,  the \emph{outneighbourhood  at time $t$} of node $v$,  ${\mathcal N}^+_t(v)$, contains time-nodes such that there exists a  directed edge  $(t,uv)$ in $E$ and its  \emph{outdegree  at time $t$} $d_t^+(v)$ is  the size of its outneighbourhood.The  \emph{inneighbourhood at time $t$} and\emph{ indegree  at time $t$} of a node are defined in the same way. We also  denote by $S(W_1,W_2)$ the substream graph of a directed stream graph $S$ induced by two time-node  subsets $W_1$ and $W_2$ of  $W$  and whose interaction subset $E_{W_1,W_2}$ is made of the interactions in $E$  from $W_1$ to $W_2$.




\section{Pattern enumeration in stream graphs}
\label{sec:algorithms}

In this section we define cores and present algorithms to compute them  and  to enumerate patterns  from (real-world) attributed stream graphs.   \subsection{Core operators}

    Let us first define  two  core operators  that will be used in our experiments in core closed pattern mining in streams.  
    We will consider as object set   the set of time-nodes $W$ of a stream graph $S=(T,V,W,E)$. 
 
     The $k$-Star-Satellite  core  operator  selects in an induced substream graph $S(W^\prime)$ high degree time-nodes together with their  neighbours and is defined through the following core property:
    \begin{definition}[$k$-Star-Satellite]
        Let $S$ be an undirected stream graph and $k\in \mathds{N}$, the  $k$-star-satellite property P$((t,v),W^\prime)$ holds  if and only if   in the induced substream graph $S(W^\prime)$  either $d_t(v) \geq k$ or there exists $(t, v^\prime)\in {\mathcal N}_t(v)$ such that $d_{t}(v^\prime) \geq k$.
        \label{def:starsat}
    \end{definition}
    The $h$-$a$ HA  core operator is a counterpart in directed stream graphs of the $h$-$a$ HA core operator in directed graphs defined  in   Sections \ref{cores}. 
    It  is designed through the  following bi-core property: 
        \begin{definition}[$h$-$a$ BHA ]
       Let $S$ be a directed stream graph and $h,a\in \mathds{N}$, the $h$-$a$ BHA property $P_b((t,v),W_1,W_2)$ holds  if and only if   in the induced substream graph $S(W_1,W_2)$, if $(v,t)$ is in $W_1$ then $d_t^+(v)\geq h$ and if  $(v,t)$ is in $W_2$ then  $d_t^-(v)\geq a$.       \label{def:hacore}
    \end{definition}
    
    The $h$-$a$ HA core of $G(X)$  is then obtained as $p(X)=H\cup A$  where $(H,A)$ is the $h$-$a$ BHA bi-core of the induced substream graph $G(X)=G(X,X)$. 
      To  define these core operators we need to prove that the associate properties are, respectively, monotone and bi-monotone properties (see Section \ref{cores}):
    
        \begin{theorem}
        Definitions~\ref{def:starsat} and~\ref{def:hacore} are respectively core and bicore properties. 
    \end{theorem}
    \begin{proof}
    
        Let us start with the $k$-Star-Satellite property~\ref{def:starsat}.
        We are interesting in proving that this property is monotonous.
        Suppose that there exists a substream $S'=(T', V', W', E')$, $S'\subseteq S$ such that for all elements $(t,v)\in W'$, property~\ref{def:starsat} holds.
        In other words, there are enough interactions in $E'$ such that node $v$ at time $t$ either has at least $k$ neighbours (and is a star), or is a neighbour of such a node (and is a satellite).
        
        Let us show that there is no stream $R=(T_R, V_R, W_R, E_R)$, $R\supset S'$ such that the property is false.
        Suppose that such a stream $R$ exists.
        Then, there exists elements of $W'$ that are not in $W_R$.
        Since the core properties defined both involves degrees, this can only mean that there are interactions in $E'$ that are not in $E_R$, which in turns means that $R\not\supset S'$.
        This validates our monotonicity claim for the $k$-Star-Satellite property.
        An identical argument can be made for Definition~\ref{def:hacore}.
    \end{proof}

\begin{figure}
        \centering
        \includegraphics[width=0.48\linewidth]{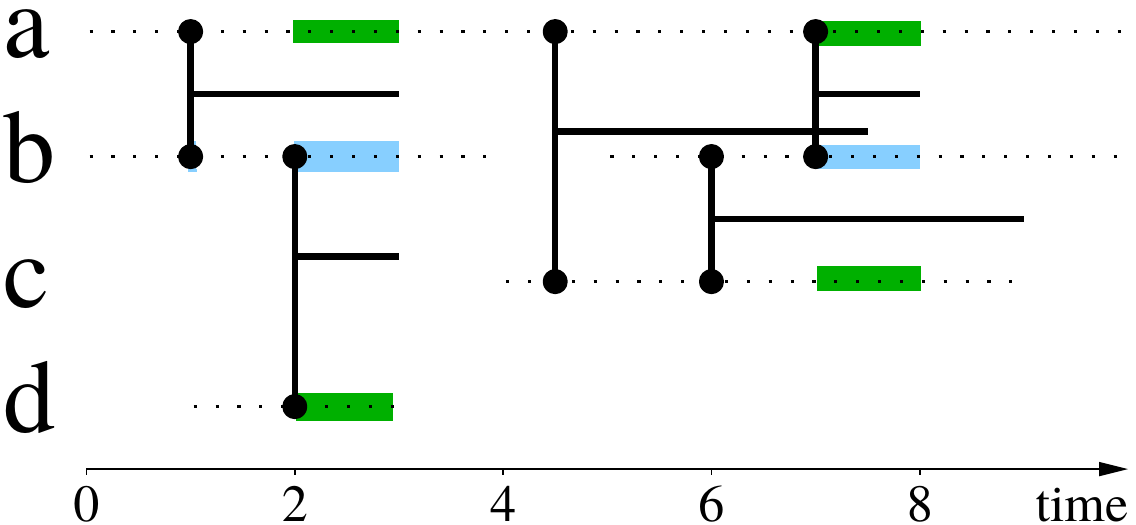}
        \includegraphics[width=0.48\linewidth]{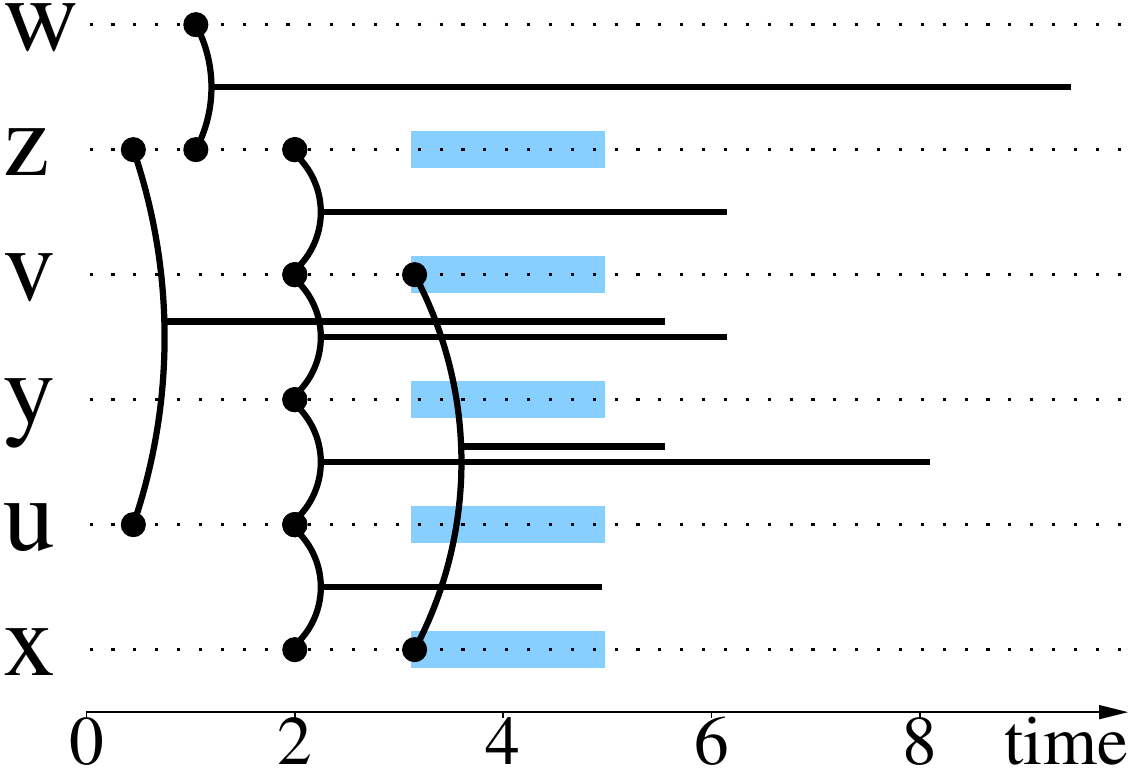}
        \caption{Illustration of the core definitions on the examples of Figure~\ref{fig:stream}. {\bf Left:} The $2$-star-satellite core, with $\{b\}\times[1,3]\cup [7,8]$ being the stars (depicted in blue), and $\{\{a\}\times [1,3]\cup[7,8], \{c\}\times [7,8], \{d\}\times [2,3]\}$ being the satellites of $b$ (depicted in green). {\bf Right:} The $2,2$-BHA-core $\{z,v,y,u,x\}\times [3,5]$. Notice that $w$ is never part of the core, since it never has degree $2$.} 
        \label{fig:cores-streams}
    \end{figure}

    Definitions \ref{def:starsat} and \ref{def:hacore} are extensions to the temporal setting of  two previously defined properties that have shown their relevance on real-world graphs~\cite{Soldano:2019aa}. 
    Figure~\ref{fig:cores-streams} illustrates these two core definitions on the toy examples of Figure~\ref{fig:stream}.
    
    \subsection{Core calculation}

Generic algorithms to compute cores are detailed in~\cite{Soldano:2019aa}. For the sake of simplicity, we consider now the single core property case. The algorithm  consider an object subset $X$  and starts a first pass in which it  remove all objects from $X$  that do not satisfy the core property $P(x,X)$, resulting in a  new $X^\prime$. A new pass is then started removing objects that do not satisfy $P(x,X^\prime)$, and the process is repeated until  a fixed point $C=p(X)$ is reached. For some properties, such as the $k$-star-satellite property, a single pass reaches the fixed point.

In stream graphs, time is modelled as continuous,  and testing for all $(t,v)$ a core or bi-core property would both (i) require some sort of discretization, (ii) result in redundant computations. Instead, note that the property is usually valid for all instants $t$ on a number of intervals of time. For instance, in Figure~\ref{fig:stream}, $(t,b)$ is a $2$-star for all $t\in [1, 2.5]$.  We obtain better algorithms by directly attempting to find the maximal such intervals. 
    As an illustration, we present Algorithm~\ref{alg:starsat}, which computes the  $k$-star-Satellite bi-core $(\mathrm{Stars},\mathrm{Sats})$ of the  substream graph $S(W)$ and returns $p(W)=\mathrm{Star} \cup \mathrm{Sats}$.       
    \input{algorithms/fastStarSat}

    Let us briefly explain our algorithm. It relies on a data structure  representing  the stream graph as a temporal adjacency table $\mathcal{D}_S$: for each node $u\in V$, we store a list $\mathcal{D}_S(u)$ of triplets $(t, v, e)$, sorted in increasing time order, indicating that node $u$ started or stopped interacting with node $v$ at time $t$.
    The flag $e\in \{1, -1\}$ indicates, respectively, the beginning or the end of an interaction.
    That structure is a discrete representation of the continuous time our object entails.
    We also rely, for all nodes $u\in V$, on a map $l_u$ that maintains, for every node, the last time when it was involved in a star or a satellite.
    
    For each interaction that appears (lines 12 to 17), we add node $v$ to the current neighbourhood ${\mathcal N}(u)$, and update $l_v$ for the current time of the interaction.
    If $|{\mathcal N}(u)|< k$ ({\em i.e.} the star property is not true), we update $l_u$ to $t$.
    Indeed, any star or satellite can only start after time $t$, since interactions are sorted temporally.
    
    Let us now focus on the case when an interaction ends (lines 18 to 27).
    First, we remove node $v$ from the current neighbourhood.
    Then we can check if the star property is valid, in which case we know that it is valid from the last relevant time for $u$, $l_u$, until now ($t$).
    We also know that all current nodes $x\in {\mathcal M}$ (the current satellites) are satellites of $u$ from time $l_u$ or $l_x$, whichever comes latest, to $t$.
    Notice that unlike in a graph, since interactions can overlap in time, it is possible for $u$ to be a star on $[t,t']$ with different satellite sets.
    

\subsection{Pattern enumeration}

Let us now discuss the pattern enumeration of all frequent core closed patterns, i.e  with core support set at least $s$.
The algorithm starts with the closure $q_0$ of the empty pattern $\emptyset$ and associated core support set $X$.
Then, for all the items $x$ ({\em i.e.} the elements of our language), we build the pattern $q_0\cup \{x\}$ and compute its core support set  in the stream, the associated core closed pattern $q_x$  and recursively all frequent core closed patterns greater than $q_x$. Maintaining a list $EL$ of  prohibited  items results in building a  a tree over the pattern lattice, in such a way that each pattern is only enumerated once. The algorithm is similar  to the one defined by~\cite{Soldano:2014cr}; indeed, thanks to the formal work presented in the previous sections, once the notions of pattern, support set  and core property are properly extended, the algorithm itself runs a similar course of execution.

\input{algorithms/bipatterns-algo}

Notice that there is a correspondence between our patterns and the ones defined in~\cite{Soldano:2014cr}. Indeed, saying that  pattern $q$ has  support set $X$  within $W$ is equivalent to saying that for any $t$, $q$ has support set $X_t= \{v \in V \mid  (t,v)\in X\}$ within $V$. In the experimental section we will consider closed patterns and core properties in the stream graph and their static counterpart in the induced graph. 




\subsection{Exhibiting patterns of interest}



Finally, let us define the  distance  to be used in the $g\beta$ selection process (see Section~\ref{sec:henry}). Given a pair of patterns $l_i$, $l_j$ and their associated core support sets $W_i, W_j$, we define their  temporal Jaccard distance as:
$$
 \sigma(l_i,l_j) = 1-   {\mathcal J}(l_i, l_j) = \frac{|W_i\cap W_j|}{|W_i\cup W_j|}
$$
$\sigma(l_i,l_j)$ has values between 0 and 1,   is equal to 0 whenever  $l_i=l_j$ and to $1$ if $l_i$ and $l_j$ have no element in common.
As a $g$ interestingness measure we consider the core support set size.

\section{Experiments}
\label{sec:results}

We now detail experiments on two real-world datasets of web and social interactions to highlight the relevance of our proposal.

    \subsection{Datasets}

\input{sections/datasets}

\input{sections/results}

\input{sections/conclusion}

%
%
%
\bibliographystyle{ACM-Reference-Format}
\bibliography{biblio,biblioH}
\end{document}

%% file: sections/intro.tex

\section{Introduction}


We consider mining connected  data with the following view:  part of the data consists in attributes values reporting information about objects, while the remaining part of the data reports information about how objects are related. We  search then for attribute patterns  i.e. sentences expressing constraints on the attributes values and  that may be valid, i.e. occur, in some objects.
 Various previous work on graphs (see Section \ref{sec:rw-cpm}) confront  such attribute patterns to the connected structure, i.e. consider poorly connected objects as poorly relevant  to the knowledge to extract. As a result the mining process enumerates and selects both attribute patterns and  the dense subgraphs associated with them.  The purpose of this article is to extend one of such methodology, namely the \emph{core closed pattern} methodology, in order to mine temporal interaction data. 

Modelling data that has a structural component over time has been done in multiple ways, and in particular recently, by considering interaction data: the connected data is then designed as a sequence of triplets $(t,u,v)$ indicating that nodes $u$ and $v$ interacted at time $t$ (see Section \ref{sec:rw-id}). They may represent, for instance, the interactions between scientists attending a conference, social networks exchanges between high school students, or interactions on the web, among others. Enriching such connection data with attributes describing individuals
allows to extract knowledge relating individuals descriptions,  to the way these individuals are  connected  at some moment. Note that the individuals descriptions may themselves depend on time: while, for instance, the background of a scientist may be considered as unrelated to the interaction time, their state of mind may depend on the time of the interaction. 

The  main characteristic of  the  \emph{stream graph} formalism is to represent interaction data is that it is based on the extensions of static graph notions in a natural way. As a consequence we may transfer conveniently results and methods from graph analysis and mining.
The present work focus on extending core closed pattern methodology to attributed stream graphs, a  process which is facilitated by the fact that the notion of graph cores, which core closed pattern mining heavily  relies on, has a natural counterpart in stream graphs.

We develop our contributions as follows: after discussing related work in Section~\ref{sec:rw}, we present the core closed pattern formalism to mine connected data in Section~\ref{sec:henry}.
In Section~\ref{sec:streams}, we present the stream graph formalism to model interactions over time, and show how to adapt the mining methodology to stream graphs.
We then present algorithms, in Section~\ref{sec:algorithms}, and apply them to closed pattern mining on two real-world datasets, in Section~\ref{sec:results}.
Finally, we conclude and present some tracks for future work in Section~\ref{sec:conclusion}.

%% file: sections/rw.tex
\section{Related work}
\label{sec:rw}

\subsection{FCA and closed pattern mining on graphs}\label{sec:rw-cpm}
A recent review on mining and finding   dense subgroups 
within attributed graphs \cite{Atzmueller:2021vo} discusses a variety of approaches,  algorithms and programs  addressing this task. Among them, various works such as \cite{Mougel2012fk}, \cite{Silva2012kx} and \cite{Soldano:2014cr} define the subgraph properties that are suitable both from formal and application standpoints. The latter introduced \emph{core closed pattern mining} whose  various definitions and results necessary  for our purpose to mine attributed interaction data are presented in Section \ref{sec:henry}. 

Closed pattern mining is strongly related to Formal concept analysis~\cite{wille2009restructuring} which  focuses on describing formally concepts associated to a \emph{context}, i.e. an object-attribute table, and ordered in a concept lattice according to a general-to-specific ordering.  
A FCA process results in producing  a lattice of concepts each made of a \emph{closed pattern}  (the concept   \emph{intent}), together with its \emph{support set} (the concept \emph{extent}) i.e.  the set of objects in which the pattern occurs. A closed pattern is then the most specific pattern among all those sharing the same support set. While FCA is a formal methodology strongly interested in the ordering of such concepts, the closed pattern mining framework focuses on the efficient enumeration of closed patterns in large datasets (see for instance \cite{Zaki:2002wi}). 

 Core closed pattern mining is a variant of closed pattern mining in which  the \emph{support set} of a pattern  is reduced to its  \emph{core support set} i.e.  the \emph{core} of the subgraph induced by the original support set. 
The first core notion is the k-core  proposed  by Seidman \cite{Seidman:1983ab} that reduces a simple and undirected graph to the unique maximal subgraph whose nodes (forming the k-core)  all have degree at least  $k$. 
Core definitions,  as generalized in \cite{Batagelj:2011fk} always rely on some topological property that have to be shared by its elements and has  proved to be a key notion for real-world network analysis. 
In  \cite{Soldano:2014cr} it is shown that the  core of a graph is  obtained by applying  an interior operator to its  vertex set, so ensuring that closed patterns exists when reducing support sets to core support sets (see Section \ref{sec:henry}). 
Adapting enumeration algorithms from closed  pattern mining~\cite{negrevergne2014miner}, that has a polynomial delay between outputting two patterns, has also been a necessary result for real-world applications.
The core closed pattern mining  framework has since then been applied to bipartite~\cite{Soldano:2019aa} and directed~\cite{Soldano:2014cr} networks, and the methodology has been extended in various ways ~\cite{Soldano:2017aa,Soldano:2017ab,atzmueller2019framework}.


\subsection{Stream graphs and modelling of interactions over time}\label{sec:rw-id}

Modelling data that has a structural component over time has been done in multiple ways, typically through different variants of dynamic graphs.
In this setting, one typically has a sequence of graphs $\{G_i\}$ and a time frame $\Delta$, and for all $i$, $E_i$ contains all the interactions that happened between times $i\Delta$ and $(i+1)\Delta$.
There are multiple variants, for example in which the graph only grows in time~\cite{george2013time}, or in which multiple concurrent values of $\Delta$ are considered~\cite{leo2019non}, but the principle remains similar.
The main limit of these approaches is linked to the loss of temporal information induced by this aggregation.
The choice of $\Delta$ is non trivial: a value too small will yield small, empty graphs, while a value too large will destroy the temporal information and the interaction causalities~\cite{caceres2013temporal}.

Recently, a few models take a different perspective, where aggregating is not necessary and one considers the sequence of interactions for itself.
The sequences of interactions are then modelled as temporal networks~\cite{holme2012temporal}, time-varying graphs~\cite{casteigts2012time} or stream graphs~\cite{latapy2018stream}, depending on the research goals and the scientific community.
In all cases, the base object is identical: a sequence of $(t,u,v)$ indicating that nodes $u$ and $v$ interacted at time $t$.
From this object, different communities have researched with different goals in mind:
temporal networks has large bodies of work around diffusion and temporal causality~\cite{holme2015modern};
time-varying graphs focuses on reachability and elaborating algorithmic complexity classes~\cite{braud2016next}; 
stream graphs focus on extending the notions used for large-graph analysis~\cite{viard2015revealing} and applying them to real-world scenarios such as traffic analysis~\cite{wilmet2019outlier}, or financial network analysis~\cite{gensollen2020you}, among others.

%% file: henry.tex

 \section{Core closed pattern mining}\label{sec:henry}

In this section we report the needed definitions and results to introduce our attributed stream graph mining methodology. Except regarding Proposition \ref{singleToBi}, they are extracted  from \cite{Soldano:2019aa}. To be self-contained, let us first recall closure and interior operator definitions: Let $S$ be an ordered set and $f: S\rightarrow S$ a self map such that for any $x,y \in S$,  $f$ is \emph{monotone}, i.e. $x \leq y$ implies $f(x) \leq f(y)$ and \emph{idempotent}, i.e. $f(f(x))=f(x)$. Then  If   $f(x) \geq x$, $f$ is called a \emph{closure operator} while  if   $f(x) \leq x$, i.e. $f$ is \emph{intensive}, $f$ is called an  \emph{interior operator}.  
 \subsection{Abstract closed  pattern mining} 
 In closed pattern mining, a pattern $q$ belongs to a pattern language $L$ which is ordered through a partial order where $q \geq q'$ means that $q$ is more specific than $q'$. Consider then  a set of objects $V$,  each object $v$ has a description $d(v)$ in  $L$  representing the most specific pattern in which it occurs, i.e.  $d(v)$ occurs in $v$ and also occurs in any pattern less specific than $d(v)$. Pattern $q$  \emph{extension},  also called its \emph{support set}, $X=\mathrm{ext}(q)$ is then the set of  its  occurrences in $V$. Applying then an interior operator 
$p$ to $\mathrm{ext}(q)$ results in reducing the support set of $q$  into its so-called \emph{abstract support set}.
 The most specific pattern  with abstract  support set $X$ is then unique, as far as the pattern language is a lattice, and is called an  \emph{abstract closed pattern}. 
 Computing the abstract closed pattern $f(q)$ with same support set as  some pattern $q$ relies on an intersection operator $\mathit{int}$ such that  $\mathit{int}(X)$ returns the most specific pattern which is less specific than    any object  description $d(o)$  in  $X$. 
We obtain then the abstract closed pattern $f(q)$ with same abstract  support set as pattern $q$, where $f$ is a closure operator, as $f(q) =\mathrm{int} \circ p \circ \mathrm{ext}(q)$.

In the closed itemset mining setting objects are described as itemsets i.e. subsets of a set of items $I$. In this case the intersection operator simply is the set theoretic intersection operator $\cap$.


\begin{example}\label{ex1}
Let us consider  $L=2^I$, $I=abcd$, $V=123$ ,$d(1)=abd,$ $d(2)=acd$, $d(3)=abc$. Pattern $\emptyset$ has support set $123$ and   $\mathrm{int}(123)=abd \cap acd \cap abc = a$.
Now consider the interior operator $p$ such that $\forall X\subseteq V, p(X)=X\setminus 3$.   We obtain then $p(123)=12$ and Following Equation \ref{two}, the abstract closed pattern $f(q)=\mathrm{int}(12)=ad$. 
\end{example}    
     
\subsection{Core closed pattern mining}       \label{cores}                                             
The following result allows us to define an interior operator on the object powerset $2^V$ from a logical property $P$ regarding an object $v$ in the context of an object subset $X$ to which it belongs:
 \begin{proposition}\label{interiorToMonotone}
Whenever a property  $P$ is monotone, i.e. for any $X\subseteq V$ and $v\in X$, we have that $P(v,X)$  and $X'\supseteq X$   implies  $P(v,X')$, then there is a unique greatest subset $C \subseteq X$  such that $P(v,C)$ holds for all $v \in C$ and  $p$ defined as  $p(X)=C$ is a an interior operator.
\end{proposition}

Using such properties is natural  when the object set  $V$ is the set of vertices of a graph $G=(V,E)$. For instance, the $k$-core \cite{Seidman:1983ab} of the subgraph $G_X$ induced by some vertex subset $X$ is defined as  the greatest  subset $C\subseteq X$ such that all vertices in $C$ have degree at least $k$ in  $G_c$, which may be rewritten as
 $P(v,C)$ holds for all $v$ in $C$. $P$ is then called a core property and  $p$ a core operator. We obtain that way abstract closed patterns, called  \emph{core closed patterns }.

A second way to obtain an interior operator on $2^V$ is  to  first  build an interior operator $p_b$ on a pair of powersets $(2^V_1,2^V_2)$ from a logical property $P_b$.  By considering then $V=V_1=V_2$ we    derive from $p_b$ a new  interior operator $p$ on $2^V$. $p_b$ is obtained as follows:
  
  \begin{proposition}\label{interiorToMonotoneBI}
  Whenever a property  $P_b$ is bi-monotone, i.e . for any $(X_1,X_2)$ pair and any $v\in X_1\cup X_2$,  $P_b(v,X_1,X_2)$  and $(X_1',X'_2) \supseteq (X_1,X_2)$   implies $P_b(v,X_1^\prime,X_2^\prime)$, then:
  \begin{itemize}
  \item there is a unique greatest subset pair $(C_1,C_2)\subseteq (X_1,X_2)$  such that $P_b(v,C_1,C_2)$ holds for all $v \in C_1 \cup C_2$ and 
   \item $p_b$ defined on $2^{V_1}\times 2^{V_2}$ as  $p_b(X_1,X_2)=(C_1,C_2)$ is a an interior operator.
\end{itemize}

\end{proposition}
\emph{Bi-cores} are then pairs of object subsets whose members  all satisfy a bi-monotone property, called a  \emph{bi-core} property.  A bi-core property $P_b$ is usually designed  from a pair of properties, i.e. $P_b(v, X_1,X_2)$ if and only if $v \in X_1$ then $P_1(v,X_1,X_2)$ holds and if $v \in X_2$  then $P_2(v,X_1,X_2)$ holds.
For instance, when $G$ is a directed graph, the $h-a$ BHA bi-core property states that in the  subgraph $G(X_1,X_2)$ induced by the directed edges from $X_1$ towards $X_2$,
if $v$ is in  $X_1$ it has  outdegree at least $h$ and if $v$ is in  $X_2$ it  has indegree at least $a$.
Note that vertices in $X_1 \cap X_2$ have to satisfy both constraints. 
    The following Proposition \ref{singleToBi} leads then to interior operators on $2^V$ and therefore to core closed patterns.\begin{proposition}\label{singleToBi}
Let $P_b$  be a bi-core  property on $(V,2^V,2^V)$  and  $p_b$ its associated  interior operator. Then, $p$ defined as 
$p(X)=X_1 \cup X_2$,  with $(X_1,X_2)= p_b(X,X)$ is an interior operator on  $2^V$ 
  \end{proposition}
  \begin{proof}
 We need to prove three properties. The proofs straightforwardly follows from the truth of the corresponding properties of the interior operator  $p_b$.
   For instance to prove that $p$ is monotone, i.e.   $X \subseteq X^\prime$ implies $p(X) \subseteq p(X^\prime)$,  we remark that  $X \subseteq X^\prime$ means  $(X,X)\subseteq (X^\prime,X^\prime)$. As $p_b$ is an interior operator this implies  $p_b(X,X)\subseteq p_b(X^\prime,X^\prime)$ and it follows that  $p(X) \subseteq p(X^\prime)$. Idempotency and intensivity are proved in the very same way.
  \end{proof}

   The $h$-$a$ hub-authority (HA)  core $p(X)$ for directed graphs  was first defined in \cite{Soldano:2017ab}.
     It may be obtained as the union of hubs $H$  and authorities\footnotemark  $A$ from the $h$-$a$ BHA bi-core  $(H,A)=p_b(X,X)$  of the  subgraph $G(X)$.   
     \footnotetext{\emph{Hub} and \emph{authoritiy}  terminology refers to the notions introduced by J M Kleinberg\cite{Kleinberg:1999aa}}  
  \subsection{Exhibiting patterns of interest}

In many real-world contexts, enumeration is only an intermediate step towards the mining of patterns of interest. When selecting individual patterns from a pattern set $Q$, according to various interestingness criteria, the resulting pattern subset  may still be redundant, i.e. contain patterns very similar to other patterns. There are various \emph{pattern set selection} ways of reducing size and redundancy of a pattern set\cite{Ouali:2017aa,Bringmann:2009aa,Vreeken:2011aa}. In our experiments we will use the $g\beta$ \emph{pattern set selection} algorithm first defined and applied to core closed patterns  in~\cite{Soldano:2019ab}.  It consists in maximizing in the selected pattern set $Q_{\beta}$ the sum of the values of a pattern interestingness measure $g$ under the constraint that two patterns $q$ and $q^\prime$ in $Q_{\beta}$ have to be at distance $\sigma(q,q^\prime)$ at least $\beta$.The  $g\beta$ pattern set selection algorithm returns a  a greedy  approximation for this problem, obtained after ordering the  input pattern list $Q$  in decreasing $g$ order.
Choosing the interestingness measure $g$, (or equivalently the corresponding pattern ordering), as well as the  distance measure $\sigma$, is typically application-dependent.

%% file: algorithms/fastStarSat.tex
\begin{algorithm}
    \caption{One-pass Stream Star Satellite algorithm}
    \begin{flushleft}
        \textbf{Input:} A  stream graph $S=(T=[\alpha, \omega],V,W,E)$   with $E$ represented as the adjacency table $\mathcal{D}_S$, a threshold $k\in \mathds{N}$ \\
        \textbf{Output:} The flattened $k$- star-satellite bi-core of $S$
    \end{flushleft}
    \begin{algorithmic}[1]
    \State Stars $\gets \emptyset$, Sats $\gets \emptyset$ \Comment{Subsets of $W$}

     \For{$u \in V$}

        \State ${\mathcal N}(u) \gets \emptyset$ \Comment{Current neighbourhood of $u$}

        \State ${\mathcal M} \gets \emptyset$ \Comment{Set of neighbours of $u$ that validate Sat property}
        \State $\forall v\in V, l_v \gets \alpha$
        \For{(v,t, e) $\in \mathcal{D}_S(u)$} \Comment{Sorted by time}
            \If{$|{\mathcal N}(u)| \geq k$}
                \State ${\mathcal M} \gets {\mathcal M} \cup {\mathcal N}(u)$
            \EndIf
            \If{e == 1}
                \State ${\mathcal N}(u) \gets {\mathcal N}(u) \cup \{v\}$
                \State $l_v \gets t$
                \If{$|{\mathcal N}(u)| < k$}
                    \State $l_u \gets t$
                \EndIf
            \Else
                \State ${\mathcal N}(u) \gets {\mathcal N}(u) \setminus \{v\}$
                \If{$|{\mathcal N}(u)| \geq k$}
                    \State Add $\{u\}\times [ l_u, t ]$ to Stars
                    \For{ $x\in {\mathcal M}$}
                        \State Add $\{x\}\times [ \max(l_x, l_u), t]$ to Sats
                    \EndFor
                    \State ${\mathcal M} \gets \emptyset$
                \EndIf
            \EndIf
        \EndFor
    \EndFor
    
    \Return Stars $\cup$ Sats

    \end{algorithmic}
    \label{alg:starsat}
\end{algorithm}

%% file: algorithms/bipatterns-algo.tex
\begin{algorithm}
    \caption{Pattern enumeration algorithm on time-node set $W$}
    \begin{algorithmic}[1]
    \State $X \gets p(W)$; 
        \State $q_0 \gets \mathrm{int}(X)$
         \State $EL \gets[] $; 
    \State \textsc{enum}$(q_0, X, EL)$

    \Function{enum}{q, X, EL}

    \State print(q, X)
    
    
    \For{$x \in I\setminus q$}
        \State $q_x \gets q \cup \{x\}$ 
        \State $X_x \gets p( \mathrm{ext}(q_x) \cap X)$
        \If{$|X_x| \geq s$}
            \State $q_x \gets \mathrm{int}(X_x)$ 

            \If{ $q_x\cap EL = \emptyset$}                   
                \State \textsc{enum}($q_x$, $X_x$, EL)
                \State $EL \gets EL \cup \{x\}$
            \EndIf
        \EndIf
    \EndFor
    
    \EndFunction
    \end{algorithmic}
    \label{alg:patterns}
\end{algorithm}

%% file: sections/datasets.tex

We performed our experiments using two data sets, one of individual contacts between high school students (HS-327), and another of research paper co-citations extracted from the Association of Computer Linguistics Anthology website.
Both datasets are publicly available, and all the code for the following experiments is available online~\footnote{\texttt{https://github.com/TiphaineV/pattern-mining}}.

\subsubsection{Contacts between individuals}

HS-327 is a dataset constructed from the results of a study of social interactions of $327$ French students conducted in 2013 \cite{mastrandrea2015plos}. The initial dataset \footnote{available for download at \url{http://www.sociopatterns.org/datasets/high-school-contact-and-friendship-networks/}} provides us with the stream of contacts over $5$ days between the students, which amounts to $33806$ temporal interactions.
The dataset also contains, for each student $u$, their class, their gender, and three lists of friends: one is the students $u$ has met (self-report), another is the students that $u$ has declared as friends (self-report), and finally, the friends $u$ has on Facebook.
We express each temporal interaction between a pair of nodes as a union of consecutive intervals of the form $[t_{i,j}-20sec, t_{i,j}]$.

\subsubsection{Academic paper citing in the ACL}

We also focus on a larger dataset. ACL-papers is built from the ACL anthology, which regroups research papers related to the Association of Computer Linguistics.
It is a co-citation temporal network, that we use to track the scientific specialities of scientists that co-author papers together between \(1979\) and \(2008\), over the span of $29$ years.
The dataset contains $250,000$ interactions between roughly $8000$ authors.
The attributes for each author in time are extracted from the abstracts' content, using the CSO ontology, as described in~\cite{Zevio:aa,Salatino:2018wd,Bird:2008vn}.
We end up with $2500$ attributes, and each author keeps all their attributes over time.
It would have been interesting to consider attributes on a per-paper basis, which we leave as future work.

%
%

%% file: sections/results.tex
\subsection{Results}

Using our implementation of the algorithms presented in Section~\ref{sec:algorithms}, we mine patterns on our two datasets.
Notice that our goal here is to showcase the potential of our method, rather than find an optimal set of parameters that will necessarily be application-dependent.

\subsubsection{HS-327}

For the HS-327 dataset, we use the $k$-star-satellite property.
We present in Table~\ref{tab:res-highschool} some numerical results depending on the value of $k$ and the selection parameter $\beta$. Notice that rapidly (when $k\geq 5$), there are no more patterns to enumerate other than the empty pattern.
This is due to the temporal nature of the data, that spreads out interactions as compared to a static graph.

\input{sections/results-hs}

In the selected patterns, we capture  generic patterns, that spread in time (for example, students of a classroom), as well as more specific patterns related to particular time intervals. This allows us to study the interactions at multiple time scales.

\begin{figure*}
    \centering
        \begin{minipage}{0.8\linewidth}
                \begin{minipage}[l][0.7\linewidth][c]{\dimexpr0.7\linewidth-0.5\Colsep\relax}
                            \includegraphics[height=0.7\linewidth]{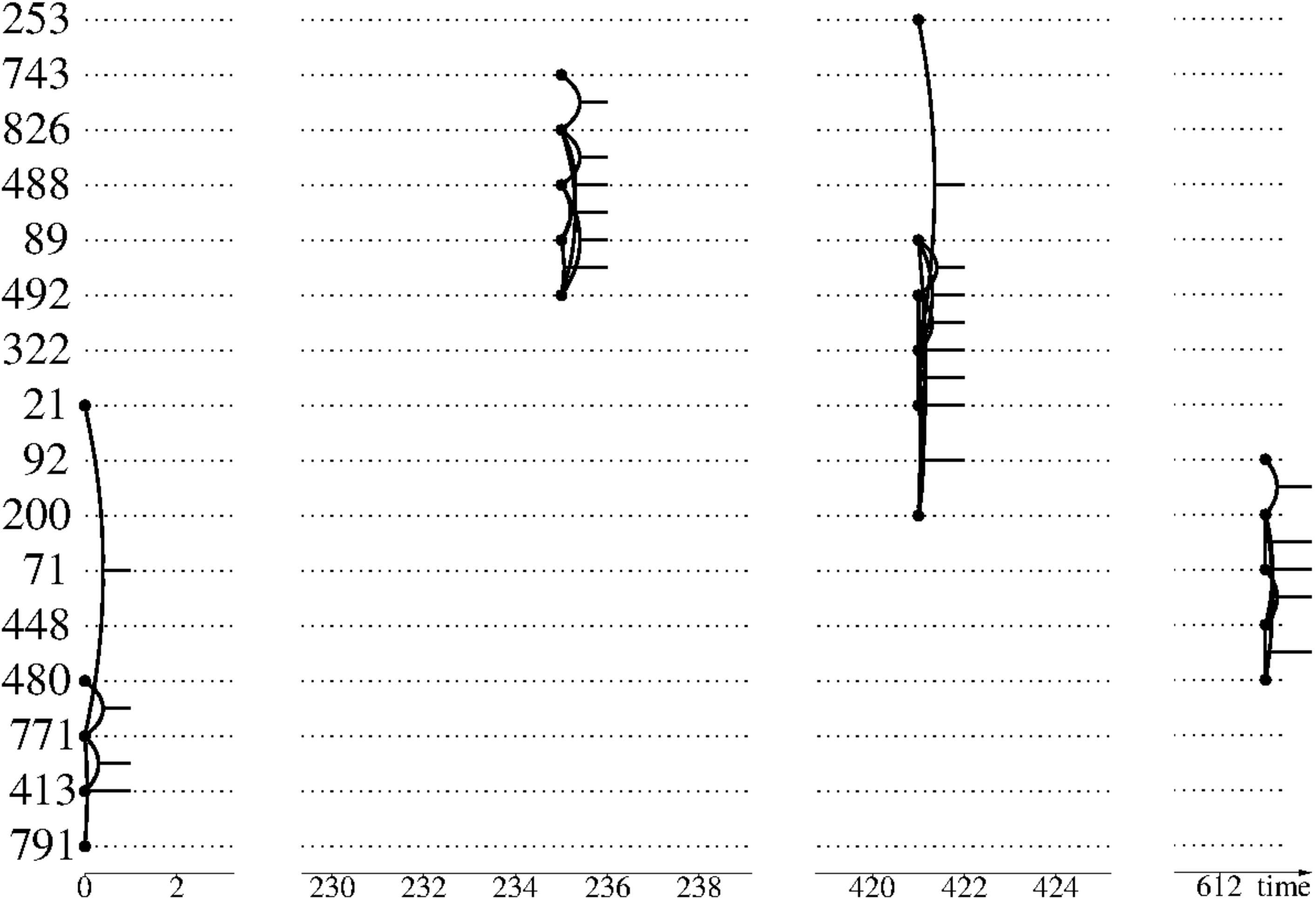}

                        \end{minipage}\hspace{2cm} 
                \begin{minipage}[l][0.3\linewidth][c]{\dimexpr0.3\linewidth-0.5\Colsep\relax}

                                \includegraphics[width=0.3\linewidth]{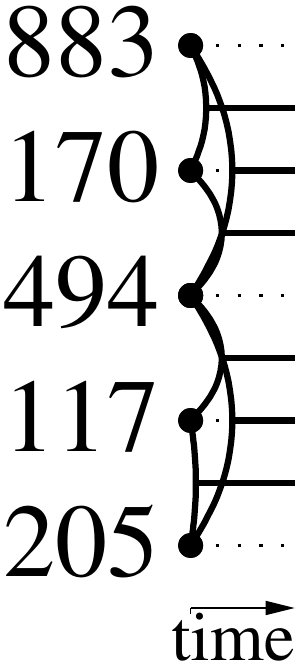} \\
                                \includegraphics[width=0.3\linewidth]{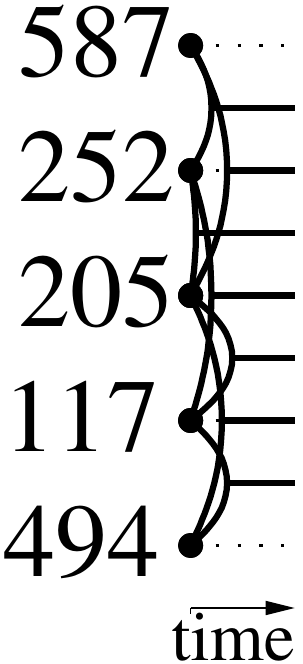} 

                \end{minipage}%
        \end{minipage}

    \caption{Some examples of patterns on the {\bf HS-327} dataset, all selected with $\beta = 0.8$. On top, one "long" pattern spread in time (blank spaces represent time periods where nothing happens). On the bottom, three more specific patterns, involving less nodes over a shorter time span. Notice that the long, less specific pattern was selected before the more specific ones.}
    \label{fig:hs-patterns-display}
\end{figure*}

As expected, a more specific pattern is correlated with smaller support sets, with the largest support set supporting the empty pattern.
However, in particular for smaller patterns, many sizes of supports sets exist.
Concerning the patterns, we noticed that many patterns contain the gender of the students (either \texttt{G\_M} or \texttt{G\_F}), reinforcing claims that students regroup in non-mixed gender groups.
In comparison, in the bottom left we display one pattern with no gender information $I$= \texttt{D\_894, F\_265, D\_205, F\_170, F\_425, F\_871, F\_1, D\_1, D\_883, F\_883, F\_205, C\_2BIO3, F\_272, F\_106}, mixing Facebook friendships and self-declared friendships.
Notice that this points to strong differences between whom the students declare as friends versus who they are Facebook friends with.
For instance, the closed pattern at bottom center is  \texttt{C\_2BIO3, D\_265, D\_272, D\_117}, this time regrouping only declared friends of the 2BIO3 class (Biology specialty). Notice that student $272$ is declared by everyone in the closed pattern as a friend, but this is not mutual.
The last pattern (bottom right) \texttt{F\_119, F\_425, F\_871, F\_1, F\_883, C\_2BIO3, F\_101} points to students that are friends on Facebook but did not declare themselves as friends.




Let us compare the patterns resulting from mining the stream to those obtained from  the static graph. To enumerate the core closed patterns from the static graph, we implemented the code from~\cite{Soldano:2019ab}.
Notice first that when considering the static graph associated to a stream graph, nodes descriptions which do not depend on time, and $k$-star-satellite cores in both cases, the core closed patterns in the stream  graph also are core closed patterns in the static graph. Indeed, if node $u$ has $k$ neighbours at a time $t\in T$, then $u$ has also $k$ neighbours in the static graph; however the converse is untrue: it is possible for $u$ to have $k$ neighbours in the static graph, each related to $u$ at different times. This means that the core definition in the static graph is a weaker constraint than the one required by the core definition in the stream graph.

\begin{figure}
    \begin{center}
     \hfill\includegraphics[height=80px]{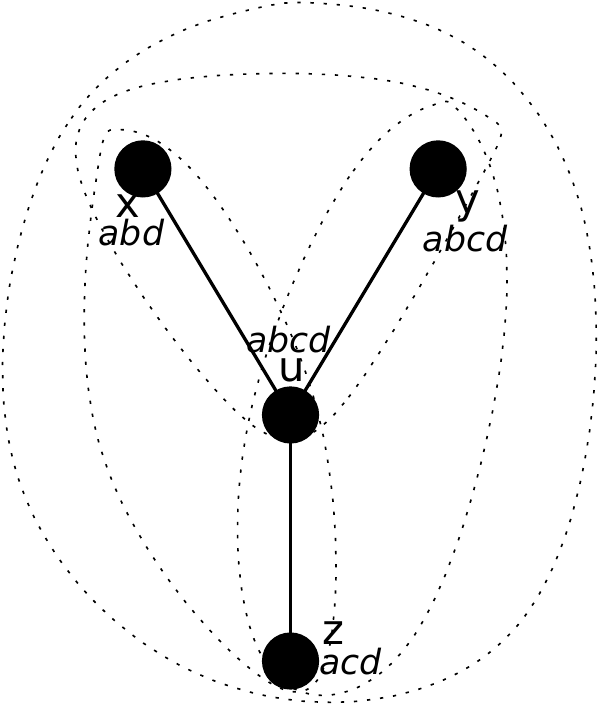} \hfill
     \includegraphics[height=80px]{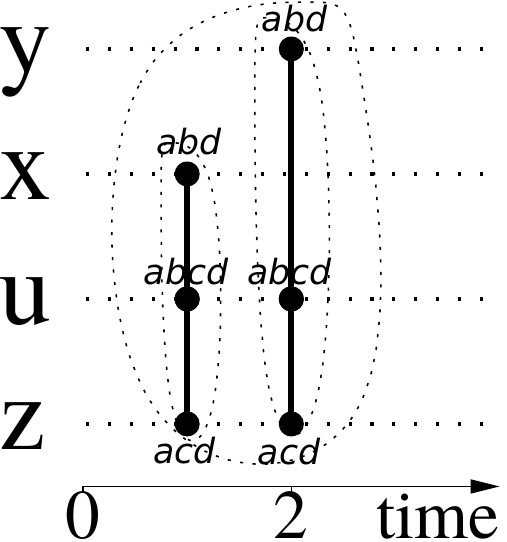} \hfill
     \caption{$2$-star-satellites on a toy stream and its induced graph. There are $4$ closed patterns on the graph, but $3$ in the stream, as the closed pattern $ab$ with core support set $\{u,x,y\}$ cannot exist in time, since $u$ never interacts with $x$ and $y$ at the same time.} 
     \label{fig:core-graph-stream}
 \end{center}
\end{figure}

As a consequence of this, when mining close patterns on the graph induced by the stream graph of the HS-327 dataset with the $k=4$-star-satellite core property, we obtain $11600$ closed patterns, to be compared to the 99 closed patterns obtained from the stream graph. Notice however that many of these patterns do not have any grounding in reality, as we show on a toy example in Figure~\ref{fig:core-graph-stream}.
In that sense, we argue that our patterns are fewer but of higher relevance.

\subsubsection{ACL}

For the ACL dataset, we mine patterns using the $h,a$-BHA-core property, and report results for $h=a=15$.
We have experimented with different values for both $h$ and $a$, and report these results since they provide enough closed patterns to be interesting without offering an overwhelming number of closed patterns.
In total, $1664$ closed patterns are enumerated in a bit less than $90$ minutes.

As for HS-327, a more specific description is correlated with fewer authors.
The intents help us highlight different subfields of the ACL Anthology; typical intents for closed patterns around 1990 involves the keywords \texttt{syntactics, context-free}, while keywords such as \texttt{learning, natural\_language\_processing} appear much later, around 2005 for most authors.

In the dataset a few ($16$) researchers are active over more than $14$ years.
This is particularly interesting, since it allows us to follow their closed patterns over time.

We can see that for most researchers, the terms \texttt{parsing} and \texttt{natural language processing} appear late (around 2003), even though one of them, \texttt{Lynette Hirschman}, has keyword \texttt{natural language understanding} in her closed patterns since 1991.
However, the support sets help paint an even more interesting picture, showing how some researchers change specialty without changing their favourite coauthors, while authors likely change domains.

Focusing on the most distinct patterns ({\em i.e.} the patterns selected with $\beta \geq 0.9$) gives other insights.
These patterns are the most mutually dissimilar according to our $g\beta$ measures.
We give the intents of these closed patterns in Figure~\ref{fig:acl-results}.
In this case, these closed patterns highlight different sub-areas of research of the Association for Computer Linguistics.
The fact that keywords co-occur even in the $0.9$-selected closed patterns likely comes from the fact that the scope of the ACL itself regroups researchers on similar topics of research.
As such, even the most dissimilar patterns retain some conceptual similarity.

\input{sections/results-acl}

%
%
%
%

%% file: sections/results-hs.tex
\begin{table}[]
\begin{tabular}{@{}llllllll@{}}
\toprule
\multirow{2}{*}{\textbf{Dataset}}       &         & \multicolumn{5}{c}{${\bm\beta}$} & \multirow{2}{*}{{\bf Runtime}}\\ \cmidrule{2-7}
               & {\bf k} & 0.0 & 0.2 & 0.4 & 0.6 & 0.8 \\ \midrule
\textbf{HS-327} & {\bf 3}      & 620 & 362 & 221 & 125 & 76  & 16mns\\
\textbf{HS-327} & {\bf 4}      & 99  & 75  & 52  & 40  & 31  & 9mns \\
\textbf{ACL} & {\bf 15, 15}      & 1030  & 406  & 175  & 56  & 12  & 90mns \\ \bottomrule
\end{tabular}
\caption{Summary of the closed patterns enumerated on both datasets, and the number of closed patterns selected by $g\beta$-selection. For the ACL dataset, we only keep closed patterns with at least $4$ keywords.}
\label{tab:res-highschool}
\end{table}

%% file: sections/results-acl.tex
\begin{figure}
{\small 
    \begin{minipage}{0.95\linewidth}

    natural languages, semantics, syntactics, 
    syntactic structure
    \smallskip
    \hrule
    \smallskip
    linguistics, machine translations, syntactics,
    syntactic structure
    \smallskip
    \hrule
    \smallskip

    bilingual, correlation analysis, machine translations, 
    translation process
    \smallskip
    \hrule
    \smallskip

    correlation analysis, learning, parsing algorithm, 
    syntactic analysis, syntactics, syntactic structure
    \smallskip
    \hrule
    \smallskip

    correlation analysis, machine translations, 
    statistical machine translation, syntactic structure
    \smallskip
    \hrule
    \smallskip

    correlation analysis, machine translations, 
    phrase-based statistical machine translation, 
    statistical machine translation, translation models

    \end{minipage}
}
    \caption{The $6$ closed patterns selected with $\beta > 0.9$ in the ACL dataset.}
    \label{fig:acl-results}
\end{figure}

%% file: sections/conclusion.tex

\section{Conclusion and Perspectives}
\label{sec:conclusion}
In this paper, we strengthen the existing bridges between formal concept analysis/closed pattern mining and real-world structural data. We show that beyond graphs, these methods can be adapted to streams of interactions, in order to mine relevant patterns from large real-world such sequences.
After recalling the notion of  {\em core} of a graph, we define two  such cores for stream graphs and show that they exhibit the necessary properties for closed pattern enumeration.
A strength of our approach is that we do not challenge the core assumptions made by previous work, allowing for little conceptual modifications algorithms from past work.
It opens the way to concurrent mining of structural data of different natures, such as a stream graph and a graph, for example.

We run experiments on two datasets, one of social, online and offline interactions between students and another based on a Web anthology of citations between scientific papers in computational linguistics. 
In both cases, we mine interesting patterns, and show that post-enumeration pattern set selection allows us to identify dissimilar patterns.

One interesting aspect of this work is the perspectives it opens, some of which we briefly detail now.
We have shown that degrees and properties around degrees offer a good trade-off between expressive power and computational efficiency; however, these properties rely on being monotone, which  limits our possibilities.
Being able to extend the scope of the theoretical framework to convex core properties would be an important progress.

This work relies on the enumeration of closed patterns to do further selection, even though only a fraction of the enumerated patterns is of interest for a typical application.
Even if we only compute a spanning tree over the concept lattice, being able to only explore sub-areas of interest is highly sought after.
This has recently be done for graphs, using local modularity~\cite{atzmueller2019framework}; there is no consensual definition of modularity for stream graphs and their variants, making this improvement an open problem. 

Application-wise, an interesting direction is the use of closed patterns to provide elements of explanation, for example as a complement to recommender systems.
One could, given a set of closed patterns and a predicted link (typically, a link between a user and a book), use the  set of closed patterns related to the user or the book to provide arguments justifying the prediction.
This would allow to tap into the growing number of resources around knowledge representation.

%% file: Exploring_mining_attributed_streams.bbl

\begin{thebibliography}{33}


\ifx \showCODEN    \undefined \def \showCODEN     #1{\unskip}     \fi
\ifx \showDOI      \undefined \def \showDOI       #1{#1}\fi
\ifx \showISBNx    \undefined \def \showISBNx     #1{\unskip}     \fi
\ifx \showISBNxiii \undefined \def \showISBNxiii  #1{\unskip}     \fi
\ifx \showISSN     \undefined \def \showISSN      #1{\unskip}     \fi
\ifx \showLCCN     \undefined \def \showLCCN      #1{\unskip}     \fi
\ifx \shownote     \undefined \def \shownote      #1{#1}          \fi
\ifx \showarticletitle \undefined \def \showarticletitle #1{#1}   \fi
\ifx \showURL      \undefined \def \showURL       {\relax}        \fi
\providecommand\bibfield[2]{#2}
\providecommand\bibinfo[2]{#2}
\providecommand\natexlab[1]{#1}
\providecommand\showeprint[2][]{arXiv:#2}

\bibitem[\protect\citeauthoryear{Atzmueller, Bloemheuvel, and
  Kloepper}{Atzmueller et~al\mbox{.}}{2019}]%
        {atzmueller2019framework}
\bibfield{author}{\bibinfo{person}{Martin Atzmueller}, \bibinfo{person}{Stefan
  Bloemheuvel}, {and} \bibinfo{person}{Benjamin Kloepper}.}
  \bibinfo{year}{2019}\natexlab{}.
\newblock \showarticletitle{A Framework for Human-Centered Exploration of
  Complex Event Log Graphs}. In \bibinfo{booktitle}{\emph{International
  Conference on Discovery Science}}. Springer, \bibinfo{pages}{335--350}.
\newblock


\bibitem[\protect\citeauthoryear{Atzmueller, G{\"u}nnemann, and
  Zimmermann}{Atzmueller et~al\mbox{.}}{2021}]%
        {Atzmueller:2021vo}
\bibfield{author}{\bibinfo{person}{Martin Atzmueller}, \bibinfo{person}{Stephan
  G{\"u}nnemann}, {and} \bibinfo{person}{Albrecht Zimmermann}.}
  \bibinfo{year}{2021}\natexlab{}.
\newblock \showarticletitle{Mining communities and their descriptions on
  attributed graphs: a survey}.
\newblock \bibinfo{journal}{\emph{Data Mining and Knowledge Discovery}}
  (\bibinfo{year}{2021}).
\newblock
\showISBNx{1573-756X}
\urldef\tempurl%
\url{https://doi.org/10.1007/s10618-021-00741-z}
\showDOI{\tempurl}


\bibitem[\protect\citeauthoryear{Batagelj and Zaversnik}{Batagelj and
  Zaversnik}{2011}]%
        {Batagelj:2011fk}
\bibfield{author}{\bibinfo{person}{Vladimir Batagelj} {and}
  \bibinfo{person}{Matjaz Zaversnik}.} \bibinfo{year}{2011}\natexlab{}.
\newblock \showarticletitle{Fast algorithms for determining (generalized) core
  groups in social networks}.
\newblock \bibinfo{journal}{\emph{Adv. Data Analysis and Classification}}
  \bibinfo{volume}{5}, \bibinfo{number}{2} (\bibinfo{year}{2011}),
  \bibinfo{pages}{129--145}.
\newblock


\bibitem[\protect\citeauthoryear{Bird, Dale, Dorr, Gibson, Joseph, Kan, Lee,
  Powley, Radev, and Tan}{Bird et~al\mbox{.}}{2008}]%
        {Bird:2008vn}
\bibfield{author}{\bibinfo{person}{Steven Bird}, \bibinfo{person}{Robert Dale},
  \bibinfo{person}{Bonnie~J. Dorr}, \bibinfo{person}{Bryan~R. Gibson},
  \bibinfo{person}{Mark~Thomas Joseph}, \bibinfo{person}{Min{-}Yen Kan},
  \bibinfo{person}{Dongwon Lee}, \bibinfo{person}{Brett Powley},
  \bibinfo{person}{Dragomir~R. Radev}, {and} \bibinfo{person}{Yee~Fan Tan}.}
  \bibinfo{year}{2008}\natexlab{}.
\newblock \showarticletitle{The {ACL} Anthology Reference Corpus: {A} Reference
  Dataset for Bibliographic Research in Computational Linguistics}. In
  \bibinfo{booktitle}{\emph{{LREC}}}. \bibinfo{publisher}{European Language
  Resources Association}.
\newblock


\bibitem[\protect\citeauthoryear{Braud-Santoni, Dubois, Kaaouachi, and
  Petit}{Braud-Santoni et~al\mbox{.}}{2016}]%
        {braud2016next}
\bibfield{author}{\bibinfo{person}{Nicolas Braud-Santoni},
  \bibinfo{person}{Swan Dubois}, \bibinfo{person}{Mohamed-Hamza Kaaouachi},
  {and} \bibinfo{person}{Franck Petit}.} \bibinfo{year}{2016}\natexlab{}.
\newblock \showarticletitle{The next 700 impossibility results in time-varying
  graphs}.
\newblock \bibinfo{journal}{\emph{International Journal of Networking and
  Computing}} \bibinfo{volume}{6}, \bibinfo{number}{1} (\bibinfo{year}{2016}),
  \bibinfo{pages}{27--41}.
\newblock


\bibitem[\protect\citeauthoryear{Bringmann and Zimmermann}{Bringmann and
  Zimmermann}{2009}]%
        {Bringmann:2009aa}
\bibfield{author}{\bibinfo{person}{Bj{\"{o}}rn Bringmann} {and}
  \bibinfo{person}{Albrecht Zimmermann}.} \bibinfo{year}{2009}\natexlab{}.
\newblock \showarticletitle{One in a million: picking the right patterns}.
\newblock \bibinfo{journal}{\emph{Knowl. Inf. Syst.}}  \bibinfo{volume}{18}
  (\bibinfo{year}{2009}).
\newblock
\urldef\tempurl%
\url{https://doi.org/10.1007/s10115-008-0136-4}
\showDOI{\tempurl}


\bibitem[\protect\citeauthoryear{Caceres and Berger-Wolf}{Caceres and
  Berger-Wolf}{2013}]%
        {caceres2013temporal}
\bibfield{author}{\bibinfo{person}{Rajmonda~Sulo Caceres} {and}
  \bibinfo{person}{Tanya Berger-Wolf}.} \bibinfo{year}{2013}\natexlab{}.
\newblock \showarticletitle{Temporal scale of dynamic networks}.
\newblock In \bibinfo{booktitle}{\emph{Temporal networks}}.
  \bibinfo{publisher}{Springer}, \bibinfo{pages}{65--94}.
\newblock


\bibitem[\protect\citeauthoryear{Casteigts, Flocchini, Quattrociocchi, and
  Santoro}{Casteigts et~al\mbox{.}}{2012}]%
        {casteigts2012time}
\bibfield{author}{\bibinfo{person}{Arnaud Casteigts}, \bibinfo{person}{Paola
  Flocchini}, \bibinfo{person}{Walter Quattrociocchi}, {and}
  \bibinfo{person}{Nicola Santoro}.} \bibinfo{year}{2012}\natexlab{}.
\newblock \showarticletitle{Time-varying graphs and dynamic networks}.
\newblock \bibinfo{journal}{\emph{International Journal of Parallel, Emergent
  and Distributed Systems}} \bibinfo{volume}{27}, \bibinfo{number}{5}
  (\bibinfo{year}{2012}), \bibinfo{pages}{387--408}.
\newblock


\bibitem[\protect\citeauthoryear{Gensollen and Latapy}{Gensollen and
  Latapy}{2020}]%
        {gensollen2020you}
\bibfield{author}{\bibinfo{person}{Nicolas Gensollen} {and}
  \bibinfo{person}{Matthieu Latapy}.} \bibinfo{year}{2020}\natexlab{}.
\newblock \showarticletitle{Do you trade with your friends or become friends
  with your trading partners? A case study in the [Formula omitted]
  cryptocurrency.}
\newblock \bibinfo{journal}{\emph{Applied Network Science}}
  \bibinfo{volume}{5}, \bibinfo{number}{1} (\bibinfo{year}{2020}),
  \bibinfo{pages}{NA--NA}.
\newblock


\bibitem[\protect\citeauthoryear{George and Kim}{George and Kim}{2013}]%
        {george2013time}
\bibfield{author}{\bibinfo{person}{Betsy George} {and} \bibinfo{person}{Sangho
  Kim}.} \bibinfo{year}{2013}\natexlab{}.
\newblock \showarticletitle{Time Aggregated Graph: A Model for Spatio-temporal
  Networks}.
\newblock In \bibinfo{booktitle}{\emph{Spatio-temporal Networks}}.
  \bibinfo{publisher}{Springer}, \bibinfo{pages}{7--24}.
\newblock


\bibitem[\protect\citeauthoryear{Holme}{Holme}{2015}]%
        {holme2015modern}
\bibfield{author}{\bibinfo{person}{Petter Holme}.}
  \bibinfo{year}{2015}\natexlab{}.
\newblock \showarticletitle{Modern temporal network theory: a colloquium}.
\newblock \bibinfo{journal}{\emph{The European Physical Journal B}}
  \bibinfo{volume}{88}, \bibinfo{number}{9} (\bibinfo{year}{2015}),
  \bibinfo{pages}{234}.
\newblock


\bibitem[\protect\citeauthoryear{Holme and Saram{\"a}ki}{Holme and
  Saram{\"a}ki}{2012}]%
        {holme2012temporal}
\bibfield{author}{\bibinfo{person}{Petter Holme} {and} \bibinfo{person}{Jari
  Saram{\"a}ki}.} \bibinfo{year}{2012}\natexlab{}.
\newblock \showarticletitle{Temporal networks}.
\newblock \bibinfo{journal}{\emph{Physics reports}} \bibinfo{volume}{519},
  \bibinfo{number}{3} (\bibinfo{year}{2012}), \bibinfo{pages}{97--125}.
\newblock


\bibitem[\protect\citeauthoryear{Kleinberg}{Kleinberg}{1999}]%
        {Kleinberg:1999aa}
\bibfield{author}{\bibinfo{person}{Jon~M Kleinberg}.}
  \bibinfo{year}{1999}\natexlab{}.
\newblock \showarticletitle{Authoritative sources in a hyperlinked
  environment}.
\newblock \bibinfo{journal}{\emph{Journal of the ACM (JACM)}}
  \bibinfo{volume}{46}, \bibinfo{number}{5} (\bibinfo{year}{1999}),
  \bibinfo{pages}{604--632}.
\newblock


\bibitem[\protect\citeauthoryear{Latapy, Viard, and Magnien}{Latapy
  et~al\mbox{.}}{2018}]%
        {latapy2018stream}
\bibfield{author}{\bibinfo{person}{Matthieu Latapy}, \bibinfo{person}{Tiphaine
  Viard}, {and} \bibinfo{person}{Cl{\'e}mence Magnien}.}
  \bibinfo{year}{2018}\natexlab{}.
\newblock \showarticletitle{Stream graphs and link streams for the modeling of
  interactions over time}.
\newblock \bibinfo{journal}{\emph{Social Network Analysis and Mining}}
  \bibinfo{volume}{8}, \bibinfo{number}{1} (\bibinfo{year}{2018}),
  \bibinfo{pages}{61}.
\newblock


\bibitem[\protect\citeauthoryear{L{\'e}o, Crespelle, and Fleury}{L{\'e}o
  et~al\mbox{.}}{2019}]%
        {leo2019non}
\bibfield{author}{\bibinfo{person}{Yannick L{\'e}o},
  \bibinfo{person}{Christophe Crespelle}, {and} \bibinfo{person}{Eric Fleury}.}
  \bibinfo{year}{2019}\natexlab{}.
\newblock \showarticletitle{Non-altering time scales for aggregation of dynamic
  networks into series of graphs}.
\newblock \bibinfo{journal}{\emph{Computer Networks}}  \bibinfo{volume}{148}
  (\bibinfo{year}{2019}), \bibinfo{pages}{108--119}.
\newblock


\bibitem[\protect\citeauthoryear{Mastrandrea, Fournet, and Barrat}{Mastrandrea
  et~al\mbox{.}}{2015}]%
        {mastrandrea2015plos}
\bibfield{author}{\bibinfo{person}{Rossana Mastrandrea}, \bibinfo{person}{Julie
  Fournet}, {and} \bibinfo{person}{Alain Barrat}.}
  \bibinfo{year}{2015}\natexlab{}.
\newblock \showarticletitle{Contact Patterns in a High School: A Comparison
  between Data Collected Using Wearable Sensors, Contact Diaries and Friendship
  Surveys}.
\newblock \bibinfo{journal}{\emph{PLOS ONE}} (\bibinfo{year}{2015}).
\newblock


\bibitem[\protect\citeauthoryear{Mougel, Rigotti, and Gandrillon}{Mougel
  et~al\mbox{.}}{2012}]%
        {Mougel2012fk}
\bibfield{author}{\bibinfo{person}{Pierre-Nicolas Mougel},
  \bibinfo{person}{Christophe Rigotti}, {and} \bibinfo{person}{Olivier
  Gandrillon}.} \bibinfo{year}{2012}\natexlab{}.
\newblock \showarticletitle{Finding Collections of k-Clique Percolated
  Components in Attributed Graphs}. In \bibinfo{booktitle}{\emph{PAKDD 2012,
  Kuala Lumpur}} \emph{(\bibinfo{series}{Lecture Notes in Computer Science})},
  Vol.~\bibinfo{volume}{7302}. \bibinfo{pages}{181--192}.
\newblock


\bibitem[\protect\citeauthoryear{Negrevergne, Termier, Rousset, and
  M{\'e}haut}{Negrevergne et~al\mbox{.}}{2014}]%
        {negrevergne2014miner}
\bibfield{author}{\bibinfo{person}{Benjamin Negrevergne},
  \bibinfo{person}{Alexandre Termier}, \bibinfo{person}{Marie-Christine
  Rousset}, {and} \bibinfo{person}{Jean-Fran{\c{c}}ois M{\'e}haut}.}
  \bibinfo{year}{2014}\natexlab{}.
\newblock \showarticletitle{Para miner: a generic pattern mining algorithm for
  multi-core architectures}.
\newblock \bibinfo{journal}{\emph{Data Mining and Knowledge Discovery}}
  (\bibinfo{year}{2014}).
\newblock


\bibitem[\protect\citeauthoryear{Ouali, Zimmermann, Loudni, Lebbah,
  Cr{\'{e}}milleux, Boizumault, and Loukil}{Ouali et~al\mbox{.}}{2017}]%
        {Ouali:2017aa}
\bibfield{author}{\bibinfo{person}{Abdelkader Ouali}, \bibinfo{person}{Albrecht
  Zimmermann}, \bibinfo{person}{Samir Loudni}, \bibinfo{person}{Yahia Lebbah},
  \bibinfo{person}{Bruno Cr{\'{e}}milleux}, \bibinfo{person}{Patrice
  Boizumault}, {and} \bibinfo{person}{Lakhdar Loukil}.}
  \bibinfo{year}{2017}\natexlab{}.
\newblock \showarticletitle{Integer Linear Programming for Pattern Set Mining;
  with an Application to Tiling}. In \bibinfo{booktitle}{\emph{PAKDD 2017,
  Jeju, South Korea, May 23-26, 2017}}.
\newblock


\bibitem[\protect\citeauthoryear{Salatino, Thanapalasingam, Mannocci, Osborne,
  and Motta}{Salatino et~al\mbox{.}}{2018}]%
        {Salatino:2018wd}
\bibfield{author}{\bibinfo{person}{Angelo Salatino}, \bibinfo{person}{Thiviyan
  Thanapalasingam}, \bibinfo{person}{Andrea Mannocci},
  \bibinfo{person}{Francesco Osborne}, {and} \bibinfo{person}{Enrico Motta}.}
  \bibinfo{year}{2018}\natexlab{}.
\newblock \showarticletitle{The Computer Science Ontology: {A} Large-Scale
  Taxonomy of Research Areas}. In \bibinfo{booktitle}{\emph{International
  Semantic Web Conference {(2)}}} \emph{(\bibinfo{series}{Lecture Notes in
  Computer Science})}. \bibinfo{pages}{187--205}.
\newblock


\bibitem[\protect\citeauthoryear{Seidman}{Seidman}{1983}]%
        {Seidman:1983ab}
\bibfield{author}{\bibinfo{person}{Stephen~B. Seidman}.}
  \bibinfo{year}{1983}\natexlab{}.
\newblock \showarticletitle{{Network structure and minimum degree}}.
\newblock \bibinfo{journal}{\emph{Social Networks}}  \bibinfo{volume}{5}
  (\bibinfo{year}{1983}), \bibinfo{pages}{269--287}.
\newblock


\bibitem[\protect\citeauthoryear{Silva, Meira, and Zaki}{Silva
  et~al\mbox{.}}{2012}]%
        {Silva2012kx}
\bibfield{author}{\bibinfo{person}{Arlei Silva}, \bibinfo{person}{Wagner Meira,
  Jr.}, {and} \bibinfo{person}{Mohammed~J. Zaki}.}
  \bibinfo{year}{2012}\natexlab{}.
\newblock \showarticletitle{Mining Attribute-structure Correlated Patterns in
  Large Attributed Graphs}.
\newblock \bibinfo{journal}{\emph{Proc. VLDB Endow.}} \bibinfo{volume}{5},
  \bibinfo{number}{5} (\bibinfo{date}{Jan.} \bibinfo{year}{2012}),
  \bibinfo{pages}{466--477}.
\newblock
\showISSN{2150-8097}


\bibitem[\protect\citeauthoryear{Soldano and Santini}{Soldano and
  Santini}{2014}]%
        {Soldano:2014cr}
\bibfield{author}{\bibinfo{person}{Henry Soldano} {and}
  \bibinfo{person}{Guillaume Santini}.} \bibinfo{year}{2014}\natexlab{}.
\newblock \showarticletitle{Graph abstraction for closed pattern mining in
  attributed networks}. In \bibinfo{booktitle}{\emph{ECAI}},
  Vol.~\bibinfo{volume}{263}.
\newblock


\bibitem[\protect\citeauthoryear{Soldano, Santini, and Bouthinon}{Soldano
  et~al\mbox{.}}{2017a}]%
        {Soldano:2017aa}
\bibfield{author}{\bibinfo{person}{Henry Soldano}, \bibinfo{person}{Guillaume
  Santini}, {and} \bibinfo{person}{Dominique Bouthinon}.}
  \bibinfo{year}{2017}\natexlab{a}.
\newblock \showarticletitle{Formal Concept Analysis of Attributed Networks}.
\newblock In \bibinfo{booktitle}{\emph{Formal Concept Analysis in Social
  Network Analysis}}, \bibfield{editor}{\bibinfo{person}{Rokia Missaoui},
  \bibinfo{person}{Sergei Obiedkov}, {and} \bibinfo{person}{Sergei Kuznetsov}}
  (Eds.). \bibinfo{publisher}{Springer}, \bibinfo{pages}{143--170}.
\newblock


\bibitem[\protect\citeauthoryear{Soldano, Santini, and Bouthinon}{Soldano
  et~al\mbox{.}}{2019a}]%
        {Soldano:2019ab}
\bibfield{author}{\bibinfo{person}{Henry Soldano}, \bibinfo{person}{Guillaume
  Santini}, {and} \bibinfo{person}{Dominique Bouthinon}.}
  \bibinfo{year}{2019}\natexlab{a}.
\newblock \showarticletitle{Attributed Graph Pattern Set Selection Under a
  Distance Constraint}. In \bibinfo{booktitle}{\emph{Complex Networks 7th
  edition, Lisbon, Portugal, December 10-12, 2019}}
  \emph{(\bibinfo{series}{Studies in Computational Intelligence})}.
  \bibinfo{publisher}{Springer}, \bibinfo{pages}{228--241}.
\newblock


\bibitem[\protect\citeauthoryear{Soldano, Santini, Bouthinon, Bary, and
  Lazega}{Soldano et~al\mbox{.}}{2019b}]%
        {Soldano:2019aa}
\bibfield{author}{\bibinfo{person}{Henry Soldano}, \bibinfo{person}{Guillaume
  Santini}, \bibinfo{person}{Dominique Bouthinon}, \bibinfo{person}{Sophie
  Bary}, {and} \bibinfo{person}{Emmanuel Lazega}.}
  \bibinfo{year}{2019}\natexlab{b}.
\newblock \showarticletitle{Bi-pattern mining of attributed networks}.
\newblock \bibinfo{journal}{\emph{Applied Network Science}}
  \bibinfo{volume}{4}, \bibinfo{number}{1} (\bibinfo{date}{6}
  \bibinfo{year}{2019}), \bibinfo{pages}{37}.
\newblock


\bibitem[\protect\citeauthoryear{Soldano, Santini, Bouthinon, and
  Lazega}{Soldano et~al\mbox{.}}{2017b}]%
        {Soldano:2017ab}
\bibfield{author}{\bibinfo{person}{Henry Soldano}, \bibinfo{person}{Guillaume
  Santini}, \bibinfo{person}{Dominique Bouthinon}, {and}
  \bibinfo{person}{Emmanuel Lazega}.} \bibinfo{year}{2017}\natexlab{b}.
\newblock \showarticletitle{Hub-Authority Cores and Attributed Directed Network
  Mining}. In \bibinfo{booktitle}{\emph{International Conference on Tools with
  Artificial Intelligence (ICTAI)}}. \bibinfo{publisher}{{IEEE} Computer
  Society}, \bibinfo{address}{Boston, MA, USA}, \bibinfo{pages}{{1120--1127}}.
\newblock


\bibitem[\protect\citeauthoryear{Viard, Latapy, and Magnien}{Viard
  et~al\mbox{.}}{2015}]%
        {viard2015revealing}
\bibfield{author}{\bibinfo{person}{Tiphaine Viard}, \bibinfo{person}{Matthieu
  Latapy}, {and} \bibinfo{person}{Cl{\'e}mence Magnien}.}
  \bibinfo{year}{2015}\natexlab{}.
\newblock \showarticletitle{Revealing contact patterns among high-school
  students using maximal cliques in link streams}. In
  \bibinfo{booktitle}{\emph{ASONAM workshop DyNo}}. IEEE,
  \bibinfo{pages}{1517--1522}.
\newblock


\bibitem[\protect\citeauthoryear{Vreeken, van Leeuwen, and Siebes}{Vreeken
  et~al\mbox{.}}{2011}]%
        {Vreeken:2011aa}
\bibfield{author}{\bibinfo{person}{Jilles Vreeken}, \bibinfo{person}{Matthijs
  van Leeuwen}, {and} \bibinfo{person}{Arno Siebes}.}
  \bibinfo{year}{2011}\natexlab{}.
\newblock \showarticletitle{Krimp: mining itemsets that compress}.
\newblock \bibinfo{journal}{\emph{Data Mining and Knowledge Discovery}}
  \bibinfo{volume}{23} (\bibinfo{year}{2011}).
\newblock


\bibitem[\protect\citeauthoryear{Wille}{Wille}{2009}]%
        {wille2009restructuring}
\bibfield{author}{\bibinfo{person}{Rudolf Wille}.}
  \bibinfo{year}{2009}\natexlab{}.
\newblock \showarticletitle{Restructuring lattice theory: an approach based on
  hierarchies of concepts}. In \bibinfo{booktitle}{\emph{International
  Conference on Formal Concept Analysis}}.
\newblock


\bibitem[\protect\citeauthoryear{Wilmet, Viard, Latapy, and
  Lamarche-Perrin}{Wilmet et~al\mbox{.}}{2019}]%
        {wilmet2019outlier}
\bibfield{author}{\bibinfo{person}{Audrey Wilmet}, \bibinfo{person}{Tiphaine
  Viard}, \bibinfo{person}{Matthieu Latapy}, {and} \bibinfo{person}{Robin
  Lamarche-Perrin}.} \bibinfo{year}{2019}\natexlab{}.
\newblock \showarticletitle{Outlier detection in IP traffic modelled as a link
  stream using the stability of degree distributions over time}.
\newblock \bibinfo{journal}{\emph{Computer Networks}}  \bibinfo{volume}{161}
  (\bibinfo{year}{2019}), \bibinfo{pages}{197--209}.
\newblock


\bibitem[\protect\citeauthoryear{Zaki and Hsiao}{Zaki and Hsiao}{2002}]%
        {Zaki:2002wi}
\bibfield{author}{\bibinfo{person}{Mohammed~Javeed Zaki} {and}
  \bibinfo{person}{Ching{-}Jiu Hsiao}.} \bibinfo{year}{2002}\natexlab{}.
\newblock \showarticletitle{{CHARM:} An Efficient Algorithm for Closed Itemset
  Mining}. In \bibinfo{booktitle}{\emph{{SDM}}}. \bibinfo{publisher}{{SIAM}},
  \bibinfo{pages}{457--473}.
\newblock


\bibitem[\protect\citeauthoryear{Zevio, Santini, Soldano, Zargayouna, and
  Charnois}{Zevio et~al\mbox{.}}{2020}]%
        {Zevio:aa}
\bibfield{author}{\bibinfo{person}{Stella Zevio}, \bibinfo{person}{Guillaume
  Santini}, \bibinfo{person}{Henry Soldano}, \bibinfo{person}{Ha{{\"\i}}fa
  Zargayouna}, {and} \bibinfo{person}{Thierry Charnois}.}
  \bibinfo{year}{2020}\natexlab{}.
\newblock \showarticletitle{A Combination of Semantic Annotation and Graph
  Mining for Expert Finding in Scholarly Data}. In
  \bibinfo{booktitle}{\emph{GEM workshop at ECML PKDD}}.
\newblock


\end{thebibliography}
